%% file: main.tex
\def\year{2021}\relax
\documentclass[letterpaper]{article} 
\usepackage{aaai21}  
\usepackage{times}  
\usepackage{helvet} 
\usepackage{courier}  
\usepackage[hyphens]{url}  
\usepackage{graphicx} 
\usepackage{natbib}  
\usepackage{caption} 
\usepackage[utf8]{inputenc} 
\usepackage[T1]{fontenc}    
\usepackage{hyperref}       
\usepackage{url}            
\usepackage{booktabs}       
\usepackage{amsfonts}       
\usepackage{nicefrac}       
\usepackage{microtype}      
\usepackage{graphicx}
\usepackage{color}
\usepackage{multirow}

\usepackage[ruled,vlined]{algorithm2e}

\newcommand{\figref}[1]{Figure~\ref{#1}}
\newcommand{\algref}[1]{Algorithm~\ref{#1}}

\newcommand{\A}{\mathcal{A}}
\newcommand{\R}{\mathcal{R}}
\renewcommand{\S}{\mathcal{S}}
\newcommand{\V}{\mathcal{V}}

\newcommand{\bbf}{{\bf f}}
\newcommand{\bbw}{{\bf w}}
\newcommand{\commentout}[1]{}

\title{Solving Sokoban with Forward-backward Reinforcement Learning}

\author{
Yaron Shoham, 
Gal Elidan
\\
}

\affiliations{
    Google research Israel\\
    yaronsh,elidan@google.com
}

\begin{document}

\maketitle

\begin{abstract}

Despite seminal advances in reinforcement learning in recent years, many domains where the rewards are sparse, e.g. given only at task completion, remain quite challenging. In such cases, it can be beneficial to tackle the task both from its beginning and end, and make the two ends meet. Existing approaches that do so, however, are not effective in the common scenario where the strategy needed near the end goal is very different from the one that is effective earlier on.

In this work we propose a novel RL approach for such settings. In short, we first train a backward-looking agent with a simple relaxed goal, and then augment the state representation of the forward-looking agent with straightforward \emph{hint features}. This allows the learned forward agent to leverage information from backward plans, without mimicking their policy. 

We demonstrate the efficacy of our approach on the challenging game of Sokoban, where we substantially surpass learned solvers that generalize across levels, and are competitive with SOTA performance of the best highly-crafted systems. Impressively, we achieve these results while learning from a small number of practice levels and using simple RL techniques.

\end{abstract}

\input{intro}
\input{related}

\input{method}
\input{sokoban}

\input{experiments}

\section{Summary}
We tackled the task of learning an effective solver agent for the challenging Sokoban puzzle where the reward is singular and given only at the goal state. 
We presented a novel approach that first solves a (possibly relaxed) reverse task, and then uses this solution to construct \emph{hint features} that help guide the forward facing agent during both training and inference time. Using our approach we were able to solve 88 of the 90 XSokoban levels, a near perfect result.
Further, to the best of our knowledge, although RL has been applied to simplistic Sokoban levels, ours is the first work that is based on machine learning and generalizes across levels to tackle the standard XSokoban benchmark set.

Importantly, using our look back before forward RL approach, we were able to achieve this state of the art result while training on a small set of just 155 practice levels, a naive linear function approximation, and standard RL techniques. Further, we demonstrated that we can achieve near state of the art results even with simple features that do not rely on an understanding of how the game should be played. We also observed that the value function learned by our method captures non-trivial strategies such as the need to push blocking boxes away from targets before executing a final packing order that allows our agent to reach its goal. 

Our work was motivated by sparse reward scenarios where we do not have access to a large number of training instances, and are thus not able to use low-level methods that are sample-heavy (e.g., \cite{Mnih+al:2015,SilverHuangEtAl16nature,Guez+al:2019}. Indeed, in our early experiments, such methods did poorly on XSokoban levels, which is not surprising given the small number of simple training levels available. In future work, it would be useful to explore a fusion of these works and our looking back approach in middle-ground scenarios where more training data is available but still not abundant.

\bibliography{sokoban}

\end{document}

%% file: intro.tex
\section{Introduction}
Coping with control and planning challenges in complex search spaces has been a core challenge of reinforcement learning (RL) for over four decades. In recent years, model free and (almost) feature free RL methods have gained wide popularity and have been able to, for example, surpass humans at playing Atari games while learning only from pixel data \cite{Mnih+al:2015} and score rewards, or defeat the world champion in the game of Go \cite{SilverHuangEtAl16nature} while learning only from board positions and the single winning reward that comes at the end of a game. Such model-free methods are not sample efficient and, more often than not, require many orders of magnitude more experiences than a human would need to reach the same proficiency. This is particularly true when the rewards are sparse (e.g., given only at the end of the game), making it difficult for the RL agent to discover favorable states.

There are many settings where we do not have access and cannot simulate a large number of training trajectories, and are thus not able to make use of such model-free approaches. In puzzle games, for example, we cannot pit one player vs. another thereby creating a de-facto infinite collection of training instances. In such cases, if we have a good model of the transition dynamics of the domain, we can sometimes use model-based RL to simulate the environment and look ahead. However, in complex settings, either the state space is too large to make such exploration practical, or we do not have access to such a transition model and training one is computationally infeasible.

\begin{figure*}[t]
    \centering
    \begin{tabular}{ccc}
    \includegraphics[width=0.45\columnwidth]{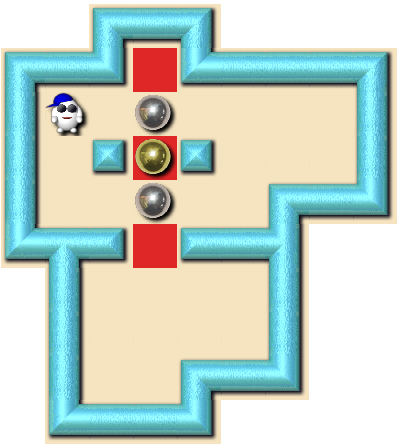} &
    \includegraphics[width=0.45\columnwidth]{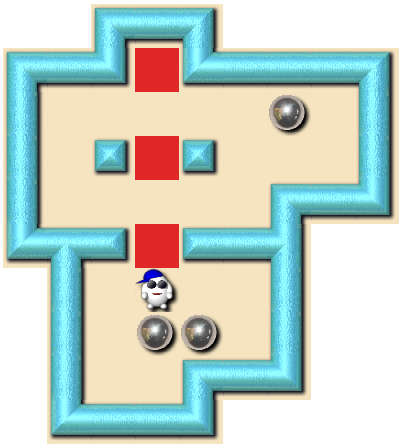} &
    \includegraphics[width=0.45\columnwidth]{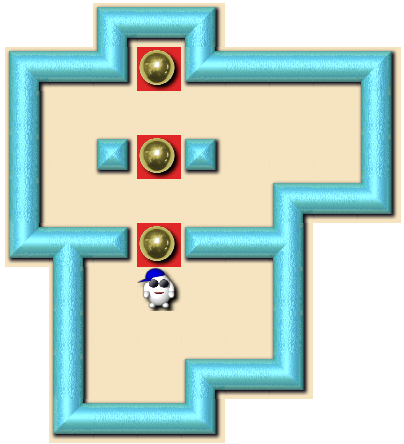} \\
    (a) Start & (b) Intermediate & (c) Goal
    \end{tabular}
    \caption{Illustration of the progression of the agent (guy with a hat) as boxes (circles) are moved in Sokoban level LOMA04-04. Note the need to first push away boxes from the targets (red squares) at the initial position (a), in order to clear a path (b) that ultimately allows the agent to pack all boxes into their final target positions (c). The strategies needed at the beginning and end phases are quite different.}
    \label{fig:SokobanExample}
\end{figure*}

In this work we focus on the challenging Japanese game of Sokoban where a player needs to push boxes into storage positions while facing the difficulty of obstacles. See \figref{fig:SokobanExample} for an example of a simple Sokoban level. Solution paths in Sokoban are much longer than in Chess or Go, the agent faces the danger of irreversible moves (e.g. when a box is pushed into a corner), and the number of relevant training levels is limited. As a consequence, Sokoban remains a game where human performance far surpasses that of automated solvers.

As in many other domains, the reward in Sokoban is given only at a single state, upon placing all boxes at their designated targets. Intuitively, the only thing we can leverage in such scenarios is the fact that the target goal(s) is known and that we can somehow backtrack from this goal in order to assist the learning of the forward looking agent. Two recent works build on this intuition and rely on the idea of data augmentation or imitation, with the inherent assumption that the agent's behavior or policy near the goal is similar to the one at the beginning of the task \cite{Edwards+al:2018,Goyal+al:2019}. Reverse curriculum learning \cite{Florensa+al:2017} suggests an interesting alternative  that gradually increases the difficulty of the learning task by considering starting states that are further and further away from the goal state. This implicitly assumes that the policy near the goal can be slowly transformed into the policy of the original problem. However, the effective policy near the goal can be quite different from the one that is beneficial earlier on. See the related work section for more details.

We suggest a simple approach that bypasses these assumptions. Instead of trying to imitate the policy of an agent that starts at the goal state and faces backwards, we use simple \emph{hint features} of trajectories of that agent to augment the state representation when learning the forward looking agent. For example, a hint-feature can be the distance to the backward plan. If this feature is minimized to zero then a solution has been found. Thus, this feature can serve as a useful signal for the forward facing agent. There are two important benefits to using these informed features instead of the policy of the backward agent directly. First, instead of imitating a specific (possibly sub-optimal) policy we allow the forward learning procedure to "discover" advantageous properties of the backward experiences via the learned value function. Second, this complete decoupling of the forward and backward policies allows us to also use relaxed backward facing agents that are given a much simpler but related goal. In the game of Sokoban, for example, the backward agent is simply required to pull all boxes away from the targets to any position on the board instead of the more difficult problem of moving them back into their original initial position. The ability to simplify the task of the backward agent can be quite important in cases where the backward task is as difficult as the forward one, and where simply walking backwards with a faraway reward will be as ineffective as walking forward.

Quite remarkably, using a small training set of just 155 beginner levels and simple RL and search algorithms, we are able to achieve state of the art results and solve 88 of the 90 benchmark XSokoban \cite{XSokoban} levels. To put this in perspective, without relying on the backward agent we solve only 60 levels, similarly to previously published works \cite{RollingStone,TalkingStones:2008}. In fact, using our looking back before looking forward RL approach, we do not only dramatically surpass other ML-based attempts but also improve the results of non-academic highly crafted systems. Importantly, as shown in the experimental evaluation section, the gains of incorporating features from the trajectories of the backward-facing agent are substantial both when using low-level features for training the forward looking agent as well as when relying on higher level ones.

Aside from the obvious contribution of making a substantial state of the art headway for the challenging game of Sokoban using learning, our contribution is in showing that by decoupling the policies of the forward and backward facing agents, we are able to make use of a simplified backward agent. This, in turn, via simple \emph{hint features}, facilitates improvement even in scenarios where the backward task is in itself quite difficult. 

%% file: related.tex
\section{Related work}
\label{sec:related}
The idea of incorporating simulated or imagined experiences to aid a learning agent is an old one and underlies the Dyna-Q algorithm \cite{DynaQ}. Prioritized sweeping \cite{PrioritizedSweeping} is an important refinement that introduces the idea of \emph{backward focusing}, that is using simulated backward trajectories to prioritize exploration in the context of model-based RL. The Queue-Dyna algorithm \cite{QueueDyna} introduces the same idea in the model free context of Q-learning. Both of these early approaches simulate backward moves from all states explored, and are not focused on faraway reward states which are hard to reach. 

Closer to our work are the more recent works of \citep{Edwards+al:2018} and \citep{Goyal+al:2019} which explicitly tackle the sparse reward scenario and initiate the backward agent at such states. Aside from various technical differences from our work such as the use of an explicit dynamics model for the backward agent, the key difference is that both approaches, either via data-augmentation or imitation, assume that the forward model can improve by mimicking the reverse policy of the backward agent. This may be true in domains that have start-end symmetries but is not generally so (see \figref{fig:SokobanExample} for an example). In contrast, we do not attempt to mimic the behavior or policy of the backward agent but instead introduce \emph{hint features} that allow the forward agent to \emph{learn} which properties of the backward trajectories are beneficial.

Many academic and practical Sokoban solvers have been developed. The milestone Rolling Stone solver \cite{RollingStone} uses a heuristic function to lower bound the number of required pushes and, using IDA* search and several domain-specific enhancements, solves 59 of the 90 XSokoban levels. The solver has no backward search component, nor does it employ learning. Talking Stones \cite{TalkingStones:2008} uses a different approach where the order in which the boxes should be packed is computed first, followed by an attempt to carry it out, and solves 54/90 levels. The packing order is reminiscent of our backward search in that working from the goal state backwards we implicitly define such an order. However, unlike our fully learned approach, their computation of the packing order is hand-crafted and works only in certain types of levels. Further, our backward facing agent produces trajectories and a packing order is just one of several features that can be extracted from these simulated experiences. 

Noteworthy non-academic solvers include Sokolution \cite{Sokolution}, Takaken \cite{Takaken} and YASS \cite{YASS} which are highly hand-crafted and solve 81, 86, and 84 levels, respectively. Solving the full set of 90 levels remained an open problem for more than 20 years until it was recently solved by the FESS algorithm \cite{FESS}. FESS relies heavily on powerful human crafted high-level features and heuristics. Our look back backward before forward approach is competitive and solves 88 levels using little training data, simple features and a fully learned value function.

The above works inherently rely on human hand crafting. More recently, DeepMind applied exciting breakthroughs in model-free reinforcement learning for Sokoban. Their Imagination-Augmented Agents (I2A) method \cite{ImaginationAugmented} performs rollouts within the evaluation function, thereby imagining the consequences of its decisions and planning ahead. Their model is able to learn the game using only low level features (pixels). Subsequent work goes a step further and shows that a general architecture (convLSTM) can exhibit many of the characteristics associated with planning \cite{Guez+al:2019}. Although these works are indeed quite innovative, in the domain of Sokoban they were only applied to simplistic artificially generated levels with few boxes (called Boxoban), and made use of a large body of $900,000$ training levels. Further, their own initial experiments indicate that performance quickly degrades as the number of boxes is increased. Our approach which uses a backward facing agent to generate \emph{hint features} and trains on a mere 155 levels, easily solves all of these simple Boxoban levels as well as 88/90 of the challenging XSokoban ones.

Another recent work that backtracks from the goal introduces the idea of \emph{reverse learning} or gradually increasing the difficulty of the learning task by pushing starting states further away from the goal \cite{Florensa+al:2017}, training an increasingly more powerful agent at each iteration. While quite appealing, the approach relies on our ability to effectively sample nearby states all the way to the original ones. This can be effective in domains where the agent cannot get stuck, e.g. robotic manipulation, but will not work in scenarios where it is easy to reach dead-ends or deadlocks, as in the game of Sokoban.

That said, the above approach can be beneficial if used in a different setup. \cite{ijcai2020-304} use this approach to quite effectively tackle difficult Sokoban levels by training a \emph{level specific} model on easier variants of that level with less boxes. The setting is quite different than ours since the learning procedure is \emph{given access to the test level} to be solved. Learning the model is thus actually part of the inference or solution process, with substantial run time ramifications: while \cite{ijcai2020-304} require 120 GPU hours to solve each new test level, our approach that generalizes across levels takes just a few minutes on a single core to solve a previously unseen puzzle.

%% file: method.tex
\section{Reinforcement Learning Preliminaries}
Reinforcement learning (RL) problems are characterized via a Markov Decision Process (MDP) that describes an environment in which actions take an agent from one state to the other, while reaping rewards or losses. $\S$ defines the collection of states $s$ in the environment. At each state, the agent can take an action $a \in \A$, and stochastically transition into another state $s'$ and receive a reward $r \in \R$, according to the transition probability $P(s',r \mid s,a)$. A policy $\pi: \S \rightarrow \A$ determines the action that the agent takes when in state $s$. Given $P$ and $\pi$, the value of each state $V(s)$, that is the expected reward when starting in $s$ and following the policy $\pi$, can be recursively computed.

Our goal in RL is to learn an effective policy, i.e., the policy $\pi^*(s)$ such that, starting from any state, the expected reward will be maximized. In value-based model-free RL, and in particular when the state space is large, instead of explicitly learning the transition function and using it to evaluate each action, we directly estimate the optimal value function $V^*(s)$ (or the action-value function $Q^*(s,a)$) from experienced or simulated agents trajectories $s_0, a_0, r_1, s_1, a_1, r_1, \ldots$, and compute the optimal policy from optimal value function. That is, the task of learning becomes the task of computing the parameters of the value function $V^*(s)$. In large state-spaces, where a tabular representation of $V(s)$ is impractical, we often use an approximate value function representation. The simplest such approximation, which is the one we use in this work, is the linear approximation $V(s) \equiv \bbw^T \bbf(s)$, where $\bbf$ is a set of features computed from the state $s$, and $\bbw$ are the model's parameters.

The approach we propose in this work can use any model-free RL method as a black-box procedure. In our application to Sokoban we use simple temporal difference, or $TD(0)$, one of the simplest yet quite effective of these methods. In $TD(0)$, given an experienced step $s_t, a_t, r_{t+1}, s_{t+1}$, the value function is updated according to the rule
\[
V(s_t) \leftarrow V(s_t) + \alpha \left[ r_{t+1} + \gamma V(s_{t+1}) - V(s_t) \right],
\]
where $\alpha$ is the learning or update rate that determines the velocity of the update rule, and $\gamma$ is the standard optional discount factor. Appealingly, this gradient-like update rule converges to the optimal value function given mild conditions and sufficient experiences or episodes (a sequence of state/action/reward that the agent goes through). However, it also highlights the difficulty in the case of sparse rewards since $r_{t+1}$ will be zero most of the time, and numerous experiences are needed for the reward to propagate backwards. See \cite{SuttonBook} for a detailed exposition of MDPs, model-free RL in general and TD learning in particular, and the concept of value function approximation. 

\section{Look Back Before Forward Reinforcement Learning}
\label{sec:method}

As discussed, our goal is to improve the learning of an agent when the signal of far-off sparse rewards is insufficient to facilitate effective learning. For simplicity of exposition we focus on the scenario where a single reward is given at the goal state, e.g., a manipulation task or a single player game. Intuitively, we can "narrow the gap" between the target and initial states by backtracking from the goal, and use this imagined experience to guide the learning process of the forward facing agent. 

However, attempting to reverse-imitate the backward policy can be detrimental. In particular, unless the domain exhibits certain regularities, the effective policy near the goal can be quite different from the one needed at the initial state. To exemplify this, consider the simple Sokoban level shown in Figure \ref{fig:SokobanExample}. Starting from the near goal position (b), an agent would learn to decrease distances between boxes and targets. However, moving from the initial state (a) to the middle state (b) requires the opposite strategy. In our experimental evaluation section we demonstrate this quantitatively.

We suggest an alternative straightforward approach where, instead of trying to mimic the trajectories of the backward agent, we build simple \emph{hint features} that relate states of these trajectories with states of the standard forward-facing agent. This allows the forward facing procedure to learn which aspects of the backward trajectories are useful for the forward policy and which are not, which in turn leads to substantial performance improvements. Below we describe the mechanics of the backward facing agent and the construction of the \emph{hint features}, the learning of the forward facing agent, and inference or the solution of an unseen task. In the next section we make this generic approach concrete for the game of Sokoban.

\subsection{The Backward Agent}
Recall that we focus on scenarios where rewards are sparse, leading to slow convergence of RL algorithms, and requiring numerous training sequences. For simplicity, and w.l.o.g., let us assume that the rewards are only given at goal states $\S_R$, and denote by $\S_I$ the set of states where the agent can be initialized (in a typical game, this will be a single state where the player starts). 

Similarly to all works that make use of backward moves (e.g. \cite{Edwards+al:2018,Florensa+al:2017,Goyal+al:2019}), we assume that the agent can be initialized at the goal states $\S_R$ and that a backward task can be executed.
With this setup in hand, the process of training the backward agent and the derivation of the backward trajectories are outlined in \algref{alg:backward} and described below.

\begin{algorithm}[t!]
\SetAlgoLined
\KwData{A set of training instances}
\KwResult{Backward value function $\V_B$} 
\BlankLine

$R_B \leftarrow$ construct backward task rewards \\
$\V_B \leftarrow$ learning using standard RL with: \\
\qquad (1) $R_B$ as reward \\
\qquad (2) Episodes initialized at goal states $S_R$ \\
\qquad (3) Reverse moves \\
\BlankLine
\Return $\V_B$
\caption{TrainBackwardModel}
\label{alg:backward}
\end{algorithm}

Our backward agent can be learned using any standard model-free RL algorithm (we use simple $TD(0)$ in this work), but starting the agent in the target positions where rewards are received $S_R$ instead of one of the initial positions $S_I$. The first step is thus to design a reversed task via the construction of reversed rewards $R_B$. We then use these rewards to train an agent while taking backward steps.

It is important to note that in some cases (e.g. Sokoban) the backward problem can be as difficult as the forward one. In such case it can be useful to present an easier task for the backward agent by using different rewards. As an example, in a large maze navigation challenge, the agent may receive a reward by simply traversing half of the (aerial) distance toward the target. As we shall see in the experimental evaluation for the  Sokoban game, trajectories of the backward agent, even when learning a relaxed problem, can be quite effective in providing useful signals to the forward agent.

\subsection{Backward Trajectories}
Having learned the value function of the backward agent $\V_B$, this model can now be used by both the forward learning procedure and the inference procedure below to generate backward trajectories from which useful \emph{hint features} will be computed. A backward trajectory $T$ for a particular (train or test) instance is created by:
\begin{enumerate}
    \item Initializing the agent at goal states $S_R$
    \item Using a planning algorithm to find $T$ that maximizes $\V_B$
\end{enumerate}
Note that we retain a trajectory $T$ even when the backward agent fails to achieve its full or relaxed goal, as even partial trajectories can be a useful guide to the forward facing agent. 

\subsection{The Forward Agent}
With the backward agent's value function $\V_B$ in hand, we are now ready to leverage this in order to construct useful \emph{hint features} and learn an effective agent whose goal is to solve the foward looking task. 

The procedure is outlined in \algref{alg:forward} and explained below. We start by construction of backward trajectories (as described above) for each training instance. We then use any standard RL algorithm to learn a forward facing value function $\V_F$ (in our experiments we use simple $TD(0)$). The agent is initialized at $S_I$ and with forward original rewards given at goal states $S_R$.  

The key difference from standard learning is that we use an augmented features space
\[
f'(s) = f(s) \cup_{i} f_i(s,T)
\]
where $f_i(s,T)$ are \emph{hint features} computed based on the current state $s$ and the instance specific backward trajectory $T$ constructed using the backward agent $\V_B$. A simple but effective example of such a feature is the Manhattan distance between the final state reached by the backward agent and the current state of the forward agent. Intuitively, such features are informed (by the backward agent) hints that can point the forward agent in the right direction, both at learning and inference time. The forward learning procedure assigns the appropriate weights to these features, in the context of the original forward task.

\begin{algorithm}[t!]
\SetAlgoLined

\KwData{Training instances and a backward model $\V_B$}
\KwResult{A forward value function $\V_F$}

\BlankLine

\ForEach{training instance $m$}{
    $T_m \leftarrow$ backward trajectory for the instance\\ \quad \qquad constructed using $\V_B$
}
\BlankLine
Use a standard RL algorithm to learn $\V_F$ with: \\
\qquad (1) Forward rewards \\
\qquad (2) Episodes initialized at initial state $S_I$ \\
\qquad (3) Forward moves \\
\qquad (4) For each instance $m$, augmentation of the\\ 
\qquad \ \ \quad features for each state encountered with \\ 
\qquad \ \ \quad  hint features $f_i(s,T_m)$
\BlankLine
\Return { $\V_F$ }
\caption{TrainForwardModel}
\label{alg:forward}
\end{algorithm}


\begin{algorithm}[t!]
\SetAlgoLined

\KwData{An instance to be solved, value functions $\V_B$ (backward agent) and $\V_F$ (forward agent)}
\KwResult{A solution to the instance}

\BlankLine

$T \leftarrow$ backward trajectory for the instance constructed using $\V_B$ \\

\BlankLine

Use a planning algorithm to maximize  $\V_F$ with: \\
\qquad (1) Start at $S_I$, forward rewards and moves\\
\qquad (2) Augmentation of features for each state\\ 
\qquad \ \ \quad encountered with hint features $f_i(s,T)$ \\

\BlankLine

\Return A solution if any of $S_G$ is reached
\caption{Inference}
\label{alg:inference}
\end{algorithm}

\subsection{Inference}
When faced with a new previously unseen task, i.e. when performing inference, we rely on both the learned backward value function $\V_B$ and the forward value function $\V_F$. The simple inference procedure is outlined in \algref{alg:inference}.

The backward agent is executed from the goal states and the resulting backward trajectory is used to construct hint features for the forward facing model. The forward agent is then applied to the original task at hand with these features and a solution is returned.
\ \\

It is important to acknowledge that this seemingly generic approach requires some specification for the domain at hand: defining the relaxed backward task; designing backward rewards; constructing hint-features. At the same time, we do note that this domain specific tailoring can be quite straightforward, as discussed in the next section for the game of Sokoban. In our experimental evaluation, we demonstrate that even with this simple tailoring we can achieve SOTA results. 

%% file: sokoban.tex
\section{Application to Sokoban}
\label{sec:sokoban}

We now make our idea of looking back before forward RL concrete for the game of Sokoban, invented by Hiroyuki Imabayashi in 1981 and published by Thinking Rabbit. The player, represented in \figref{fig:SokobanExample} as a guy with a hat, can move around in all four directions and is able to push boxes one square ahead, if that space is vacant. The player cannot pull or push a box sideways. The goal in each level is to push all boxes to target positions, depicted as red squares. Most levels are meant to be challenging, creative and fun, and humans can solve them in a reasonable time. Theoretically, Sokoban has been shown to be NP-hard \cite{SokobanNP}, and in practice it still poses a great challenge to automated solvers: it has a large branching factor, requires long solutions, and is riddled with irreversible moves that result in deadlocked states, e.g. when a box is pushed into a corner. We start by describing the value function representation, followed by a short exposition of the specifics of the forward and backward agents.

\begin{figure*}[t!]
    \centering
    \begin{tabular}{cccc}
    \includegraphics[width=0.5\columnwidth]{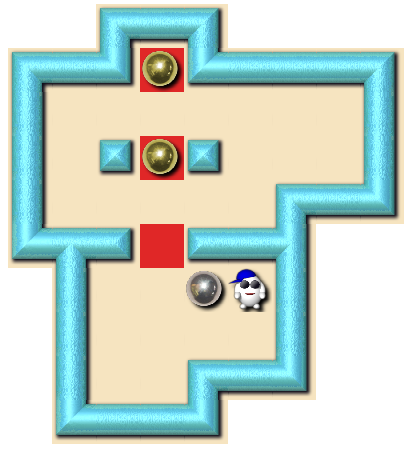} &
    \includegraphics[width=0.5\columnwidth]{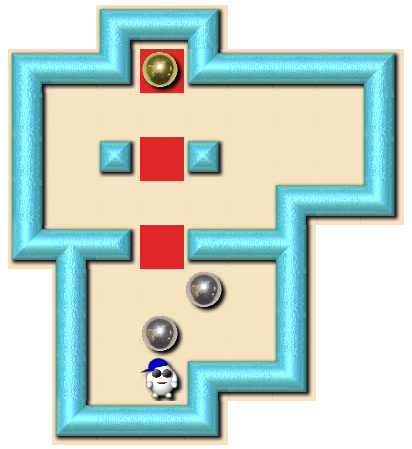} &
    \includegraphics[width=0.5\columnwidth]{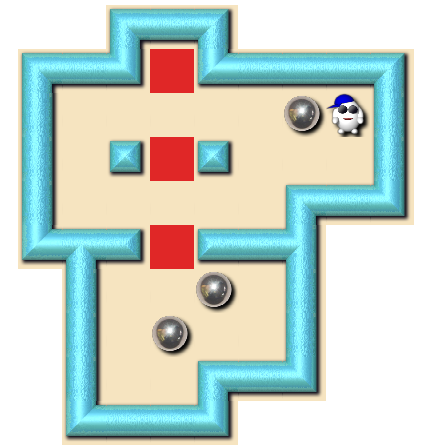} &
    \includegraphics[width=0.5\columnwidth]{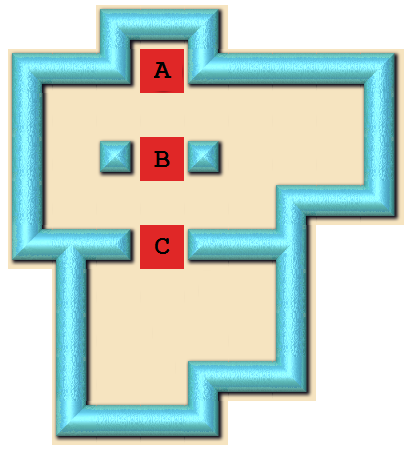} \\
    (a) After a single pull & (b) Two pulls & (c) Three pulls & (d) The packing order implied
    \end{tabular}
    \caption{Illustration of the progress of the backward facing agent as it pulls boxes away from targets. (a) shows the position after pulling the bottom box away from the target and clearing the path for the agent to pull the middle box (b). Finally, the agent can clear the top box (c) and a top-middle-bottom packing order is implied (d).}
    \label{fig:TrajectoriesExample}
\end{figure*}

\subsection{Value Function Representation}
As discussed, we use the $TD(0)$ learning algorithm with a simple linear approximation for the representation of the value function. The core features used by both the forward and backward agents are:
\begin{itemize}
\item {\bf Targets}: a count of the number of boxes already packed on target squares.
\item {\bf Distance}: the total number of pushes required to move all boxes onto target squares. This feature does not take into account the order in which the boxes should be pushed nor the fact that intermediate moves that push boxes away from targets may be required. Therefore, it is a lower bound on the actual number of pushes needed.
\item {\bf Gamma1}: A crude estimate of the reward for solving the level, while taking reward discounting into account. Concretely, $\gamma_1 = \gamma ^ {\#boxes}$.
\item {\bf Gamma2}: A refined version of the Gamma1 estimate, taking into account the number of boxes that have already been packed. Concretely, $\gamma_2 = \gamma ^ {\#boxes - \#packed}$. For the backward search, the number of packed boxes is replaced by the number of unpacked ones.
\end{itemize}
The above features are all quite low-level in that they are only based on simple knowledge of the game rules and objective (packing of boxes), without any insight into how the game should be played. As we show in the experimental evaluation, these features are all that is needed to reach results that are quite impressive and that are already competitive with state-of-the-art performance. In our final solver we also use the following higher level feature:
\begin{itemize}
\item {\bf Connectivity}: Sokoban positions can be divided into regions. In each region, the player can move freely. The player, however,  cannot move to another region without first pushing boxes that block the path that connects the regions. For example, in \figref{fig:SokobanExample} (c) the connectivity is 2.
\end{itemize}
All of the above features are normalized to the range $[0,1]$. For example, one of five boxes on a target will result in a {\bf Targets} value of 0.2. This normalization allows the model to generalize across levels of different difficulty, and keeps the gradients stable during training. Note that, despite this, the number of boxes in a level does affect the expected reward via the discounting factor $\gamma$.

\subsection{Looking Backward Before Forward}
The backward agent is initialized at the goal state, i.e., in a board where all the boxes are packed in target positions. The backward agent is trained to solve a relaxed problem and receives a reward as soon as \emph{all} boxes are pulled away from the target squares to \emph{anywhere} on the board, even if they are not back at their original positions. Note that we do not use any reward shaping, and rely solely on the simplest possible reward. Indeed,  tailoring of the backward task and its rewards requires only a rudimentary understanding of the puzzle.

Before describing the specifics of how we use the backward trajectories in the context of Sokoban, it is instructive to see an example progress of the backward agent as it starts in the final position and attempts to pull boxes away from targets as shown in \figref{fig:TrajectoriesExample}. Note that even positions (a) and (b), before the agent has managed to complete the task, reveal some configurations that could be useful hints for the forward search. Thus, such intermediate positions can also provide useful \emph{hint features} that will point the forward facing agent in the right direction. Indeed, in complex levels our backward agent does not always succeed yet its trajectories still allow us to substantially improve the forward performance.

Given the trajectory of the backward agent, we augment the features set with two additional \emph{hint features}:
\begin{itemize}
    \item {\bf Overlap}. A simple way of measuring similarity between board positions is to count the number of boxes which are on the same squares. The overlap feature is the maximum of this number measured between the current state $s$ and all states of the backward trajectory $s_b \in T$. This is analogous to measuring the distance between a point and a curve. Note that the approach is general and does not rely on any Sokoban specific knowledge.

    \item {\bf Perm}. This is a higher level feature that counts the number of boxes in the forward state $s$ that are on target squares, but measured according to the packing order implied by the backward trajectory. In this trajectory, boxes are removed from targets in a specific order. Reversing this plan suggests an order in which boxes should be packed. See figure 2(d) for an example of a packing order.

\end{itemize}

All the features that we use are taken from existing Sokoban literature, with no further feature-engineering. The only exception is the {\bf Overlap} hint-feature, which relates the backward trajectory to the forward search and is unique to our method. We do note, however, that this is a simplistic feature which only counts boxes as a naive measure of the distance between board positions. It does not contain any further inductive bias as to how the puzzle should be tackled.

\subsection{Training Setup}

{\bf Training set}. We train both the forward and backward agents using only the 155 Microban levels \cite{Microban} which have been developed as "a good set for beginners and children". The forward facing agent receives a reward of 1 when all boxes are on targets, and the backward facing agent receives a reward of 1 when it achieves the easier goal of unpacking all boxes away from the targets to any position on the board. 

{\bf Search Tree}. During both training and inference, as is standard when using RL for puzzle solving, a search tree is maintained. At each iteration a node in the search tree is expanded and its children are assigned a value based on the current value function and a reward, if received. The value of the best move serves as a new, better estimate for the position chosen for expansion, and the model weights are updated, as in standard $TD(0)$. 
The planning algorithm that we use for node expansion is a standard $\epsilon$-greedy search: traversal of the tree begins at the root and descent is to the highest value child with probability $1-\epsilon$, and uniformly over children otherwise. When a leaf is reached, it is expanded. Note that the learning is off-policy: the value function is updated using the best move, even when not picked by the $\epsilon$-greedy policy.

When the construction of the backward search tree is complete, we pick the leaf with the highest value. The branch from the root to the leaf determines the backward trajectory.

{\bf Training epochs}. In each iteration, the 155 training levels are randomly permuted and searched by the agent until a reward is received, with a cap of 50, 100 search nodes for the backward and forward agents, respectively.
We perform 100 training iterations starting with a $TD(0)$ learning rate $\alpha$=0.01, and multiplying it by $0.98$ at each iteration. Training takes a few minutes on a single core.

%% file: experiments.tex
\section{Experimental Evaluation}
\label{sec:experiments}

We now evaluate the merit of our look back before forward reinforcement learning approach using \emph{hint features} on the challenging game of Sokoban.  

\subsection{Experimental Setup} 
As outlined in the previous section, we train on the simple 155 Microban levels \cite{Microban}. We evaluate performance on the 90 XSokoban levels \cite{XSokoban} which present a medium-level challenge to humans but that are still quite challenging to automated solvers.

As previously described (see \algref{alg:inference}), given trained backward and forward agents, each test level is processed at inference time as follows. We first let the backward agent tackle it, starting from the goal state. We then use the trajectory of that agent to compute our \emph{hint features}, and then apply the forward facing agent with these features. The exact mechanics of the agents are identical to the learning phase and described in the training setup part of the previous section. The backward search is capped at 10000 tree nodes, and the forward search at 50000, which corresponds to ~5 minutes on a single core machine, with no special optimization.

We compare our solver to a baseline that is trained in an identical manner but without the \emph{hint features} that are generated by the backward agent. We also compare to a baseline based on reverse curriculum learning \cite{Florensa+al:2017}, adopted to Sokoban: we pull away n/2 boxes from the target, resulting in positions that are close to the target but not degenerate. We then augment the training data with simpler tasks by adding instances with these positions as the starting points.

Baselines for other Sokoban solvers are taken from the Sokoban solvers statistics site \cite{SokobanSolversSite}, which compares various solvers on several data sets. The site considers a level to be solved if a solution is found in less than 10 minutes, and our experiments meet this criteria.

\subsection{XSokoban Results}
\label{XSokoban_results}

\begin{figure}[t!]
\centering
\includegraphics[scale=0.3]{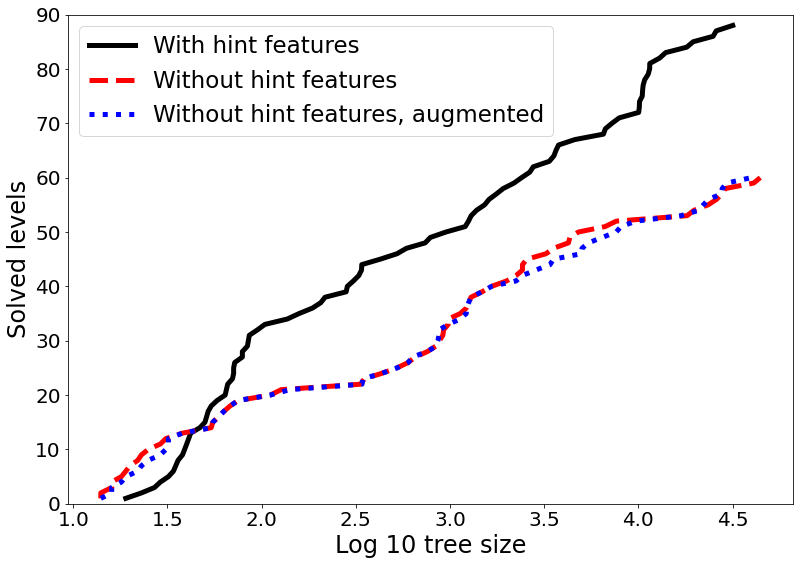}
\caption{Summary performance on the 90 XSokoban test levels. The graph shows the number of solved levels as a function of the search tree size at inference time. Compared are our method (solid black), a baseline without \emph{hint features} (dashed red), and the baseline augmented with near-goal trajectories (dotted blue). }
\label{fig:XSokoban_results}
\end {figure}

The results are summarized in \figref{fig:XSokoban_results}. Using our look backward before forward approach (solid black line), we are able to solve 88 of the 90 XSokoban levels. This surpasses the highly crafted solvers Sokolution \cite{Sokolution}, Takaken \cite{Takaken} and YASS \cite{YASS} that can only solve 81, 86, 84 levels, respectively. Our result is second only to the FESS solver \cite{FESS}. Impressively, we are able to do so while training on only 155 simple levels and using straightforward RL techniques.

The added value of the \emph{hint features} is obvious and without them we are only able to solve 60 levels. Further, augmenting the experiences of the forward solver with near-goal trajectories leads only to a minor improvement. This is due to the fact that the backward and forward setting can be quite different, emphasizing the importance of \emph{deriving} features from the backward agent rather than trying to mimic it.

\begin{table}[t!]
\begin{center}

\begin{tabular}{l | c | c | c | c |}
\cline{2-5}
& Overlap & Perm & Connect & Solved \\
\hline 
\multirow{2}{*}{{\bf No Hints}} & no & no & no & 55\\
\cline{2-5}
& no & no & yes & 60\\  
\hline \hline
\multirow{4}{*}{{\bf With Hints}} & yes & no & no & 82\\
\cline{2-5}
& yes & no & yes & 86\\  
\cline{2-5}
& yes & yes & no & 86\\
\cline{2-5}
& {\bf yes} & {\bf yes} & {\bf yes} & {\bf 88} \\
\hline

\end{tabular}
\caption{Summary performance on the 90 test XSokoban levels for different combination of features. Shown is the number of levels solved for each variant. Note the
substantial gap between the standard {\bf No Hints} learning setup and our {\bf With Hints} approach.}
\label{table:features}

\end{center}
\end{table}

It is natural to wonder how the effectiveness of the \emph{hint features} depends on the quality of the features used. Specifically, whether higher level features such as {\bf Connectivity} or {\bf Perm} that arise from an understanding of the game are needed for our approach to be useful. To evaluate this, we also tried learning forward and backward agents with only the most basic features, i.e. without connectivity and using only the overlap count as a \emph{hint feature}. As can be seen in Table \ref{table:features}, even the weakest look back before looking forward variant is able to solve 82 levels, already putting it on par with the best existing solvers, highlighting the power of \emph{hint features} even in simple settings.

\begin{figure}[t]
\centering
\includegraphics[scale=0.3]{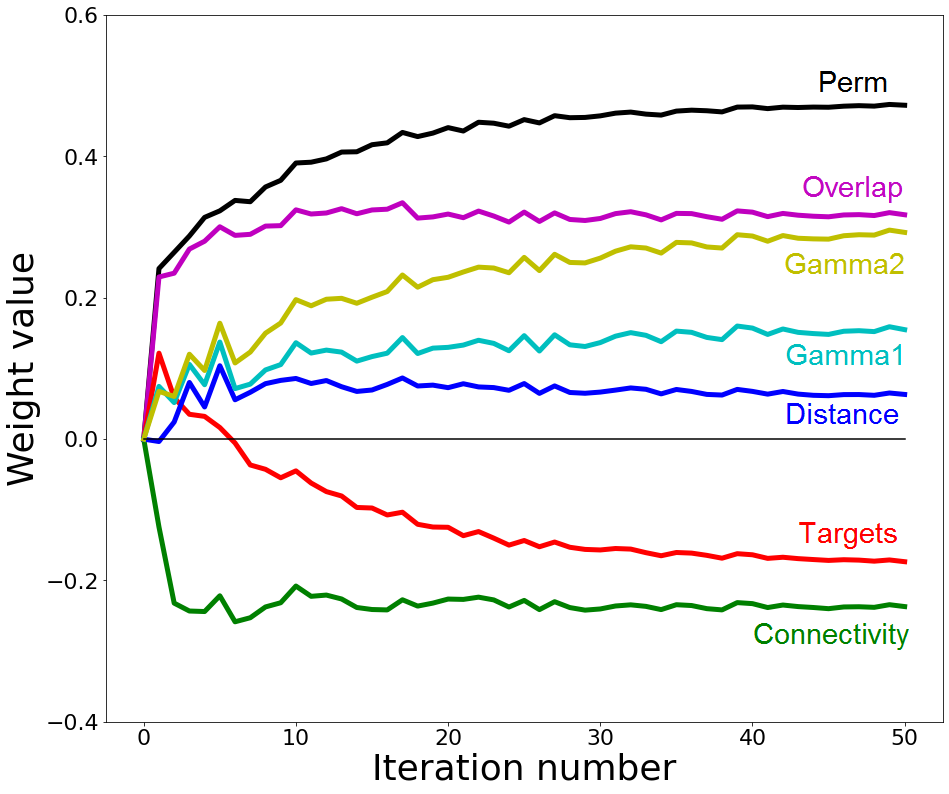}
\caption{The weights of the linear value function $\V_F$ as it progresses during the learning iterations of our looking back before forward approach. } 
\label{fig:weights}
\end {figure}

Finally, it is insightful to look at the weights of the linear approximation value function. These are shown in \figref{fig:weights}, as they progress during learning iterations. 
The weight of {\bf Connectivity} (green) is negative, as expected. The weight of {\bf Targets} (red) starts out as positive as the agent gets its first rewards. The agent then learns that moving boxes to targets is detrimental if this does not follow the packing plan, and {\bf Targets} drifts to a negative value. The positivity of {\bf Distance} (blue) might be surprising for those who have not played Sokoban but indeed one often needs to first push boxes away from the targets in order to clear the path for the final packing, as is the case in the example of \figref{fig:SokobanExample}. Most appealingly, as might be expected from the quantitative results above, the \emph{hint features} {\bf Perm} (black) and {\bf Overlap} (purple) play a dominant role.